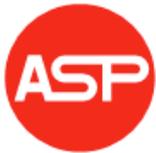

# Synergistic Integration of Techniques of VC, Communication Technologies and Unities of Calculation Transportable for Generate a System Embedded That Monitors Pyroclastic Flows in Real Time.

Barrera Kevin[1,2], Cruz Christyan[1,2], Viteri Xavier[1,2], Ing. Mendoza Darío[1,2]

[1]Mechatronics Engineering Degree, Universidad de las Fuerzas Armadas ESPE Extension Latacunga, Cotopaxi, Ecuador
[2]Robotics laboratory, Universidad de las Fuerzas Armadas ESPE Extensión Latacunga, Cotopaxi, Ecuador

At the end of an extensive investigation of the volcanic eruptions in the world, we determined patterns that coincide in this process, this data can be analyzed by artificial vision, obtaining the largest amount of information from images in an embedded system, using monitoring algorithms for compare continuous matrices, control camera positioning and link this information with mass communication technologies. The present work shows the development of a viable early warning technology solution that allows to analyze the behavior of volcanic flows automatically in a rash in real time, with a very high level of efficiency in the analysis of possible trajectories, direction and quantity of the lava flows as well as the massive mass media directed to the affected people.

Keywords: Arduino, Open CV, Artificial Vision, Volcanoes.

## 1. INTRODUCTION

Through thousands of generations, the human being has been affected by catastrophic events that have reduced the population of the same, being one of these events the volcanic eruptions of great impact, since the humanity is in constant struggle to survive these events, for which they have been in need of implementing methods for the constant monitoring of volcanoes. Since ancient times, when a person was designated to night watch if a volcano changed state[1], until the present day where different techniques for volcanic monitoring are being implemented, the focus of the present work essentially is based on the processing of images[2]. A system was developed which is in charge of monitoring in real time flows of lava, molten rock that is expelled from a reservoir close to the earth's surface during a volcanic eruption.

*Email address: kibarrera@espe.edu.ec

In 2015 in the province of Cotopaxi, Ecuador was put in a state of emergency due to the possible reactivation of the volcano in this area, causing panic among the inhabitants of the villages, for that reason, focused on using technological tools to benefit the Society to persevere human lives this system was developed, being of great importance so that it can give early alerts to the population, the system will be able to operate at all times and thus generate information in what direction the lava flows are going, in order to try to reduce the number of victims.

When using VC (computer vision) in embedded systems we can obtain a greater precision in the acquisition of data in real time[3,4], that is to improve the technique of visual observation in volcanoes, analyzing each image and thus to determine disturbances or anomalies between pixels, finally recognizing digitized volcanic flows by linking them with scheduling algorithms to different early warning systems[5].

## 2. BIBLIOGRAPHICAL REVIEW

Lava flows are the amount of molten rock ejected from the cone of a volcano, sliding down the slope of the volcano; there are four types of flows, these are:
1. In blocks.
2. Aa (Stones with rough lava)
3. Pahoehoe (liquid and soft)
4. Padding (stone forms under the sea)

The Autonomous Popular University of the State of Puebla in Mexico, is the pioneer for the development of a satellite monitoring system for the Popocatépetl volcano[6], its main difficulty is the construction of the Nano satellite to follow the volcano, thus demonstrating the need to integrate Science and technology for the benefit of society. The main difference with our system is the possibility to determine existing disturbances by a continuous comparison of layers and in turn identify the affected populations in that eruption, these being alerted to the instant of the occurrence of a variation in the volcanic cone or the pyroclastic flows.

The Geophysical Institute of Ecuador (GIE) constantly monitors active and potentially active volcanoes in Ecuador. These observatories are classified from the surveillance level one to three according to the personnel and monitoring resources they possess, instruments such as electronic inclinometers and infrared airborne cameras are used by a person in charge to verify the state of the volcano every day of the year, for this reason it was proposed to create a system that allows to optimize resources and to use the components that GIE already has the most efficient, making all observatories in the country level three[7], which means constant remote monitoring of volcanoes.

## 3. METHODOLOGY

Because the operator of the infrared cameras currently has to work long hours which affects his health, it was necessary to develop a system in which the operator will only have to take care of the maintenance of the equipment for the analysis by artificial vision. Using variables that are intertwined in this research as the amount of flow, velocity, position, natural factors, environmental factors, perturbations and a type of camera[8,9], it was necessary to investigate each of these sections, to propose the following method.
a. Segment images of volcanoes
b. Compare variations between matrices.
c. Filter the disturbances on each pixel.
d. Calculate the existing displacement.
e. Generate an angle of motion.
f. Send data via GSM and HTML

The operation of this research was verified by the implementation of a model starting with a simple analysis, continuing with the visualization of real volcanoes in the country using data from the GSE cameras located in strategic monitoring sites in Ecuador and ending with the analysis of eruptions recorded with cameras common throughout the world of which it can be affirmed that the system has a reliability of 99.4%, as long as the cameras used are optimal for the means of application.

Since this is an exploratory or pilot study since there are very few studies on the application of artificial vision in volcanoes[10,11], it was necessary to get technical advice with the different authorities in charge of the monitoring of the volcanoes in Ecuador, the hypothetic-deductive method was used, analyzing the problem of a population, through deductive reasoning validating our hypothesis empirically for the project that can safeguard human lives.

## 4. DEVELOPMENT

### 4.1 ARTIFICIAL VISION IN OPEN CV

Using Open CV in Windows and Android, images are acquired from cameras located in strategic locations to have a complete visualization of the volcano, arrays are processed and analyzed to understand a volcanic eruption, analysis were performed in Qt Creator and Android Studio, which are in charge of executable and transportable applications.

### 4.1.1 Image processing

From the f/1.4-5.6 system cameras, it is necessary to obtain the largest amount of image information with a resolution of 2560x1440 pixels, segmenting the images into HSV color spaces with Hue between 139-202 and RGB, makes a comparison of the captured image at the instant and the previous one to this event possible, thus creating the comparison of perturbations by layers, Fig.1. Describes said process.

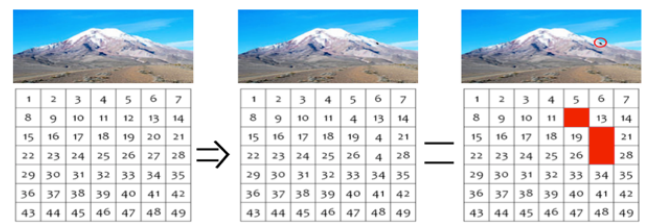

Fig.1. Comparison of perturbation by layers

The program is designed to automatically record video of the data obtained prior to an adjustment to the image and in the case that an eruption is detected, an image of the event is stored in a previously specified route using the functions.

    1 VideoWriter grab() y capture.set();

Once the perturbation is obtained, different filters and layers are used so that the analysis can be the desired one.
1. cvtColor();inRange();
2. morphOps();
3. trackFilteredObject();
4. erodeElement =

```
   getStructuringElement( MORPH_RECT,Size(2,
   1));
5  dilateElement =
   getStructuringElement( MORPH_RECT,Size(4,
   2));
```

### 4.1.2 Determination of trajectories

In the creation of the edges and position vectors PCA main component analysis and area detection are used to determine each of the different positions of the flows with respect to the camera, with values obtained using the circle function and the morphs function, a comparison is made in each of the layers using polar, rectangular coordinates and the Hough transformed, Fig.2. The application of this function to the volcanic cone can be seen in Fig.3. to finally indicate the number of objects, along with their area, position and trajectory.

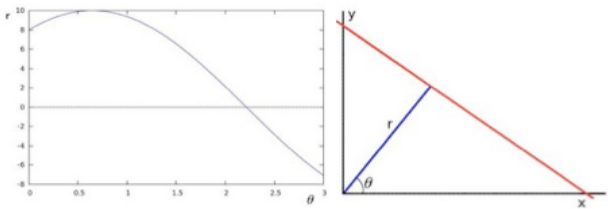

Fig.2. Hough Transform

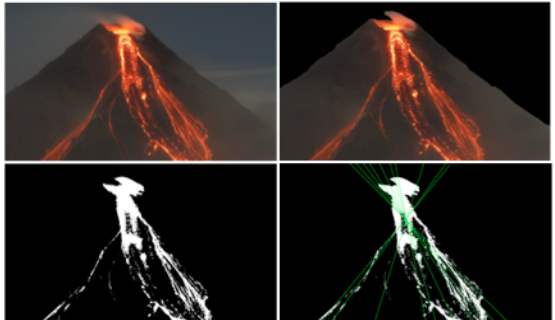

Fig.3. Application of Hough Transform in volcanoes

This principle draws a straight line according to the values obtained by PCA, the parametric space that is formed between the two points of the line creates an angle which is clear from theorem 1.

$$y = \left(-\frac{cos\theta}{sin\theta}\right)x + \left(\frac{r}{sin\theta}\right)$$

THEOREM 1. Equation Angulo Hough

Before printing the results, the program asks for the angle of the vector that is generated in the object, comparing it in a database to finally determine the direction of the flow.

1. if(grados>90 &&
   grados<180){putText(img,"Direccion Sur-Oeste: ",pt7,1.75,1.2,Scalar(0,0,255),2);
2. std::string varAsString = std::to_string(abs(grados-45));
   putText(img,varAsString,pt8,1.75,1.2,Scalar(0,0,0),2);}

For the final result the values are directed to a conditional structure, which asks if one or more flows have been generated, and it is here when the values are printed on the screen and the values are sent by serial connection to the Arduino which is in charge of generating an answer with each of these values.

## 4.2 ARDUINO MONITORING

### 4.2.1 Remote Monitoring by HTML

The Arduino receives the serial values and interprets them to perform different functions, it is connected to a LAN with its own web page in HTML programming, where remote monitoring of volcanoes can be done, using a web address, in where we create:

1. The mac address: byte mac[] = { 0xDE, 0xAD, 0xBE, 0xEF, 0xFE, 0xED };
2. Server IP: byte ip[] = { 192, 168, 0, 120};
3. IP of the web page: byte gateway[] = { 192, 168, 0, 1 };
4. Submask: byte subnet[] = { 255, 255, 255, 0 };
5. The communication port: EthernetServer server(80);

This system has been designed with 40-180 bits encryption and user access and password security. Finally, the result before the different users of the remote page will be:

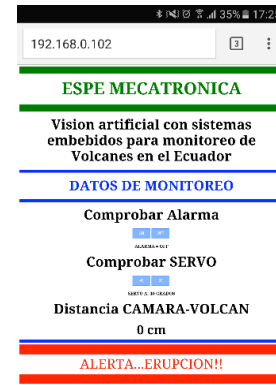

Fig.4. HTML Page in Arduino

### 4.2.2 Serial Local Monitoring via Bluetooth

From Open CV data is sent by a COM port, in the Arduino used an HC05 which is the bluetooth module to receive and send such data, it is necessary that the user previously paired the bluetooth device, the code in Arduino for this implementation is:

1. char word = Serial.read();

To determine what function the Arduino does the "word" is sent locally from a PC, a mobile device or from the HTML page, these values go to "if conditional" cycle for the different interactions with the monitoring.

## 4.3 MONITORING FROM THE ANDROID PLATFORM

Nowadays most people have a Smartphone which is very useful for this monitoring system, all these values are sent serially from Open CV to Arduino and this in turn uploads the values to a cloud in which to enter it is necessary to use this application. Thanks to the possibility of compiling open CV libraries in Android studio, it was possible to perform all the comparison of disturbances from a Smartphone, making this technology become transportable with an air monitoring from a drone.

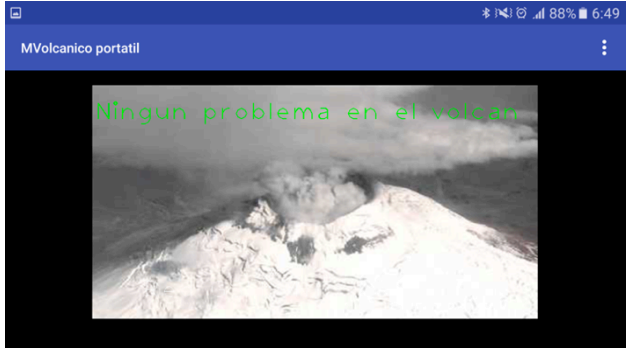

Fig.5. Overflying Cotopaxi volcano Ecuador, using portable Android application

### 4.4 INTERFACE QT CREATOR AND ANDROID STUDIO

To make the executable file in Qt, it is necessary to have an operator-friendly interface, not only so that he can visualize the volcano in addition to the program itself, without any kind of intervention, it is necessary to have a degree of security for it. Which was created a user authentication and password.

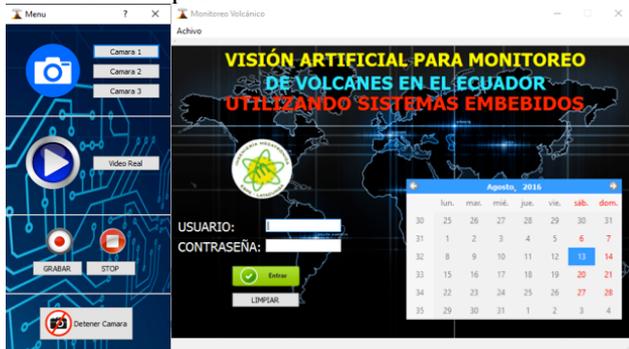

Fig.6. File .exe interface PC

In the transportable application there are two modes, the user mode will immediately receive the notification corresponding to the eruption, in case of one, this notification generates advices for users about the eruption, such as the direction of flows, Algorithm is designed to transmit the data by the cellular radiotelephony system type GSM, to receive text messages from a cellular company, all this focused on the possible populations in emergency.

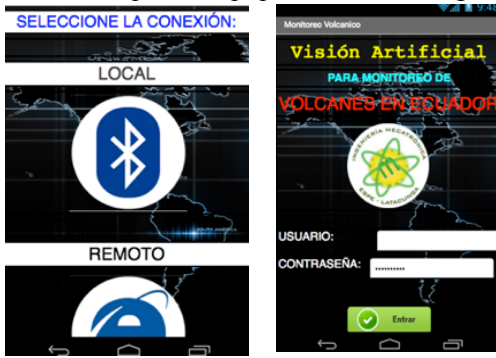

Fig.7. Android application interface

## 5 EXPERIMENTAL RESULTS

When applying this system in a constant monitoring the following results were obtained based on the variation of area, perimeter, type of eruption, visibility of the cameras and widening of the volcanic cone, see Table 1.

Table.1. Result, test volcanic eruptions

| TESTS | | DETECTION |
|---|---|---|
| 1 | 350T | 100% |
| 2 | 100T | 100% |
| 3 | 250T | 100% |
| 4 | 590T | 98% |
| 5 | 160T | 100% |
| 6 | 480T | 100% |
| 7 | 560T | 99% |
| 8 | 420T | 100% |
| 9 | 650T | 97% |
| 10 | 360T | 100% |
| Result | Succ | 99.40% |
| | Error | 0.60% |

The result of the data can be visualized in the Fig.8. which reflects the percentage of the operation of the system.

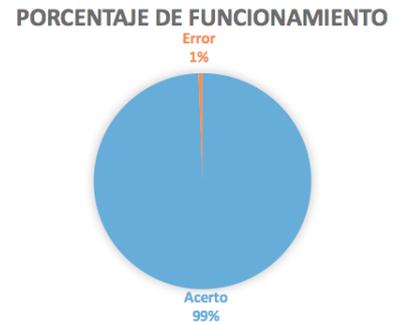

Fig.8. Percentage Operating Result

Finally, for the test of results achieved, a compendium of 52 videos of volcanic eruptions was visualized, the system worked as expected, with the following results, as we can see in the Fig.9,10,11,12.

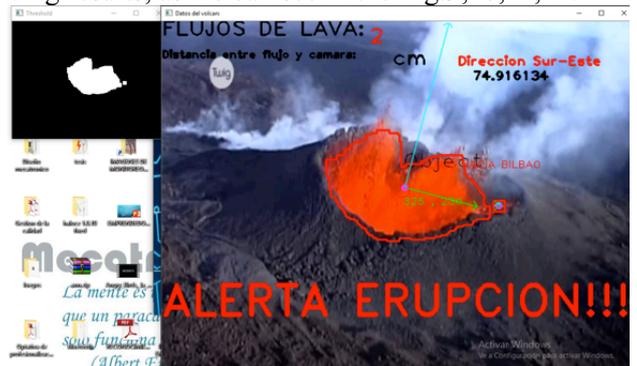

Fig.9. Analysis in active volcanic crater, aerial shot

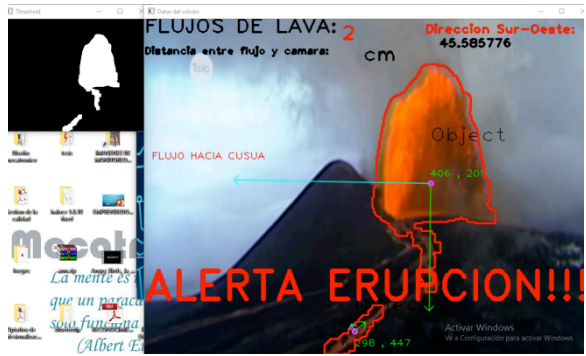

Fig.10. Static camera analysis at 900m from the volcano

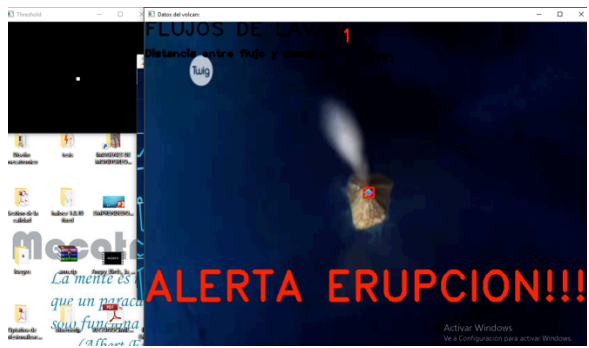

Fig.11. Analysis from satellite camera

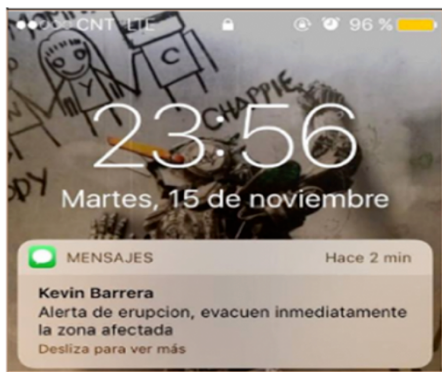

Fig.12. Notification of volcanic eruption

## 6. CONCLUSIONS

Based on the results obtained the program helps to have a permanent monitoring every day of the year with a great field of accuracy, since it determines each type of flow, the area it presents, it's possible trajectories depending on the behavior of the volcano, enclosing the area and determining the exact positioning of the center of the same with respect to the camera, in addition to the different operating systems and alert both the common user and the operator of this program since a person can be monitoring the volcano from the comfort of one's home thanks to the HTML page and likewise the people affected in the eruption can anticipate any kind of catastrophe by visualizing where the lava flows, saving a greater number of lives by using early warnings, it is necessary to have a camera that can acquire an image without interference of noise and that works effectively as the cameras of the Geophysical Institute of Ecuador, this system will help less lives to be lost in future disasters of nature with the use of new technologies in a field that has not yet been explored in depth.

## ACKNOWLEDGMENTS

This work is supported by the Arm Force University ESPE Extension Latacunga.